\documentclass[10pt,twocolumn,letterpaper]{article}
\pdfoutput = 1
\usepackage{cvpr}
\usepackage{times}
\usepackage{epsfig}
\usepackage{graphicx}
\usepackage{amsmath}
\usepackage{amssymb}
\usepackage{amsthm}
\usepackage{caption}
\usepackage{subcaption}
\usepackage{url}
\usepackage{makecell}

\usepackage[pagebackref=true,breaklinks=true,letterpaper=true,colorlinks,bookmarks=false]{hyperref}

\cvprfinalcopy 

\def\cvprPaperID{266} 

\newcommand\blfootnote[1]{%
  \begingroup
  \renewcommand\thefootnote{}\footnote{#1}%
  \addtocounter{footnote}{-1}%
  \endgroup
}

\ifcvprfinal\fi
\begin{document}

\title{Deep3DShape: 3D ShapeNetswork Joint 3D Recognition and Volume Completion}
\title{Convolutional 3D ShapeNets for Joint Recognition and Reconstruction}
\title{Deep 3D Net }
\title{3D ShapeNets for Joint Object Recognition and Reconstruction}
\title{Convolutional Shape Net for 3D Recognition and Reconstruction}
\title{3D ShapeNets for Joint Object Reconstruction and Recognition}

\title{3D ShapeNets for Joint Object Reconstruction and Recognition}

\title{3D ShapeNets for Next-Best-View in Object Recognition}
\title{3D ShapeNets for Next-Best-View RGB-D Recognition}
\title{3D ShapeNets for Next-Best-View Object Recognition}
\title{3D ShapeNets for Object Recognition and Next-Best-View Determination}
\title{3D ShapeNets for Object Recognition and Next-Best-View Planning}
\title{3D ShapeNets for Next-Best-View Planning\\ in Object Recognition}
\title{3D ShapeNets for Next-Best-View Planning\\ in 2.5D Object Recognition}
\title{3D ShapeNets for 2.5D Object Recognition\\ and Next-Best-View Planning}
\title{3D ShapeNets for 2.5D Object Recognition and Next-Best-View Prediction}
\title{3D ShapeNets: A Deep Representation for Volumetric Shapes}

\author{\normalsize 
Zhirong Wu$^{\dagger\star}$~~~~Shuran Song$^\dagger$~~~~Aditya Khosla$^\ddagger$~~~~Fisher Yu$^\dagger$~~~~Linguang Zhang$^\dagger$~~~~Xiaoou Tang$^\star$~~~~Jianxiong Xiao$^\dagger$
\\\normalsize$^\dagger$Princeton University~~~~~~~$^\star$Chinese University of Hong Kong~~~~~~~$^\ddagger$Massachusetts Institute of Technology}

\maketitle
\vspace*{-6mm}

\begin{abstract}
3D shape is a crucial but heavily underutilized cue in today's computer vision systems,
mostly due to the lack of a good generic shape representation.
With the recent availability of inexpensive 2.5D depth sensors (e.g. Microsoft Kinect),
it is becoming increasingly important to have a powerful 3D shape representation in the loop.
Apart from category recognition, recovering full 3D shapes from view-based 2.5D depth maps is also a critical part of visual understanding. 
To this end, we propose to represent a geometric 3D shape as a probability distribution of binary variables on a 3D voxel grid, using a Convolutional Deep Belief Network. 
Our model, 3D ShapeNets, learns the distribution of complex 3D shapes across different object categories and arbitrary poses from raw CAD data,
and discovers hierarchical compositional part representations automatically. 
It naturally supports joint object recognition and shape completion from 2.5D depth maps, and 
it enables active object recognition through view planning.
To train our 3D deep learning model, 
we construct ModelNet -- a large-scale 3D CAD model dataset.
Extensive experiments show that our 3D deep representation enables significant performance improvement over the-state-of-the-arts in a variety of tasks.
\blfootnote{$^{\dagger}$This work was done when Zhirong Wu was a VSRC visiting student at Princeton University.}
\end{abstract}

\section{Introduction}

\begin{figure}[t]
\centering

\includegraphics[width=1\linewidth]{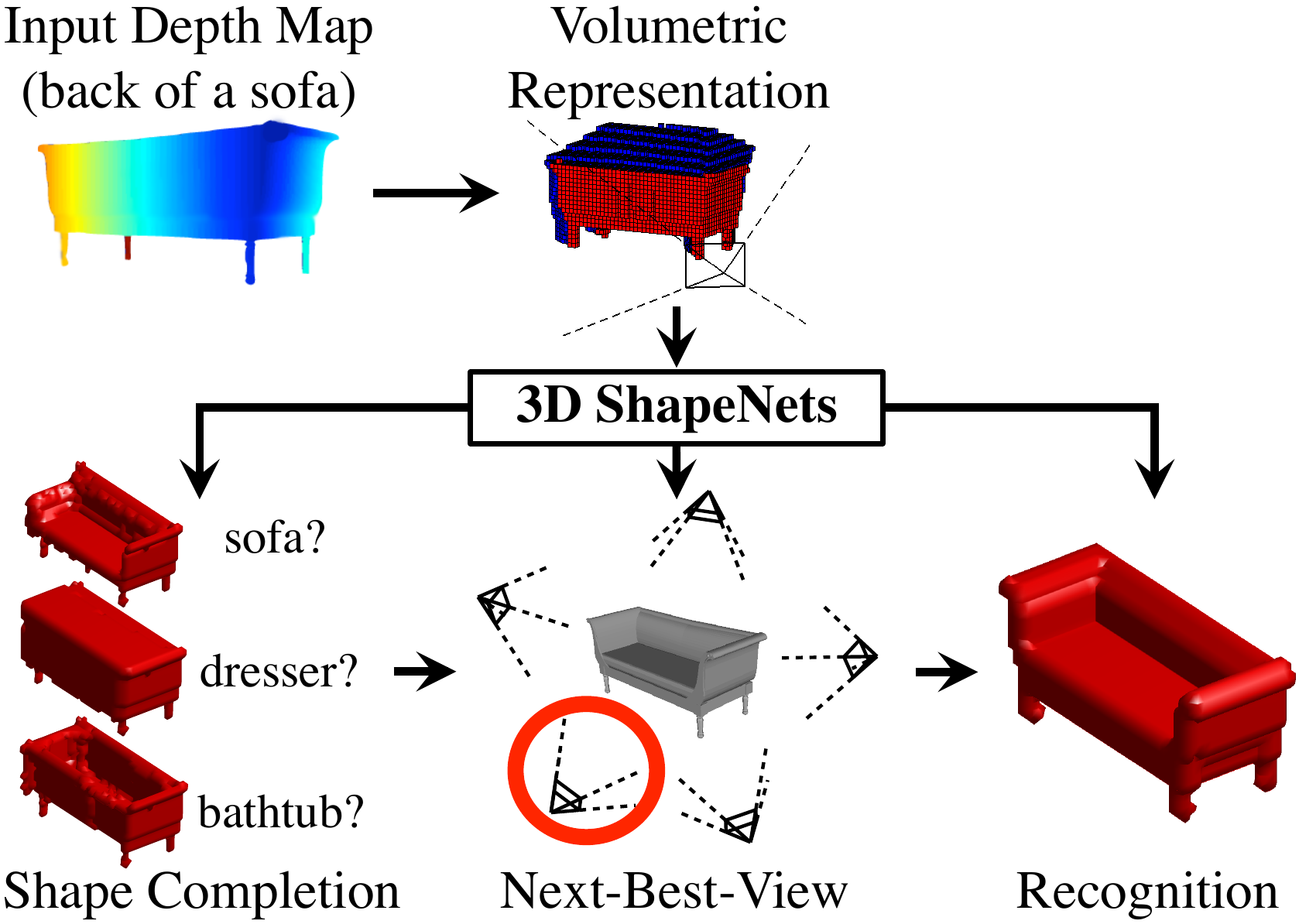}

\normalsize\href{http://3DShapeNets.cs.princeton.edu}{http://3DShapeNets.cs.princeton.edu}

\vspace{-1mm}
\caption{{\bf Usages of 3D ShapeNets.}
Given a depth map of an object, we convert it into a volumetric representation and identify the observed surface, free space and occluded space. 
3D ShapeNets can recognize object category, complete full 3D shape, 
and predict the next best view if the initial recognition is uncertain. 
Finally, 3D ShapeNets can integrate new views to recognize object jointly with all views.
}

\label{fig:teaser}
\end{figure}

Since the establishment of computer vision as a field five decades ago,
3D geometric shape has been considered to be one of the most important cues in object recognition.
Even though there are many theories about 3D representation (e.g. \cite{geon,GeometricEra}),
the success of 3D-based methods has largely been limited to instance recognition
(e.g. 
model-based 
keypoint matching to nearest neighbors \cite{rothganger20063d,tang2012textured}).
For object category recognition, 3D shape is not used in any state-of-the-art recognition methods (e.g. \cite{DPM,DCNN}),
mostly due to the lack of a good generic representation for 3D geometric shapes.
%
%
Furthermore, the recent availability of inexpensive 2.5D depth sensors,
such as the Microsoft Kinect, Intel RealSense, Google Project Tango, and Apple PrimeSense,
has led to a renewed interest in 2.5D object recognition from depth maps (e.g. Sliding Shapes \cite{SlidingShapes}).
Because the depth from these sensors is very reliable, 3D shape can play a more important role in a recognition pipeline.
As a result, it is becoming increasingly important to have a strong 3D shape representation in modern computer vision systems.

Apart from category recognition,
another natural and challenging task for recognition is shape completion:
given a 2.5D depth map of an object from one view, what are the possible 3D structures behind it? For example, humans do not need to see the legs of a table to know that they are there and potentially what they might look like behind the visible surface. Similarly, even though we may see a coffee mug from its side, we know that it would have empty space in the middle, and a handle on the side.

\begin{figure*}
   \centering
	\begin{subfigure}{0.27\textwidth}
			\centering
             \includegraphics[width=1\textwidth]{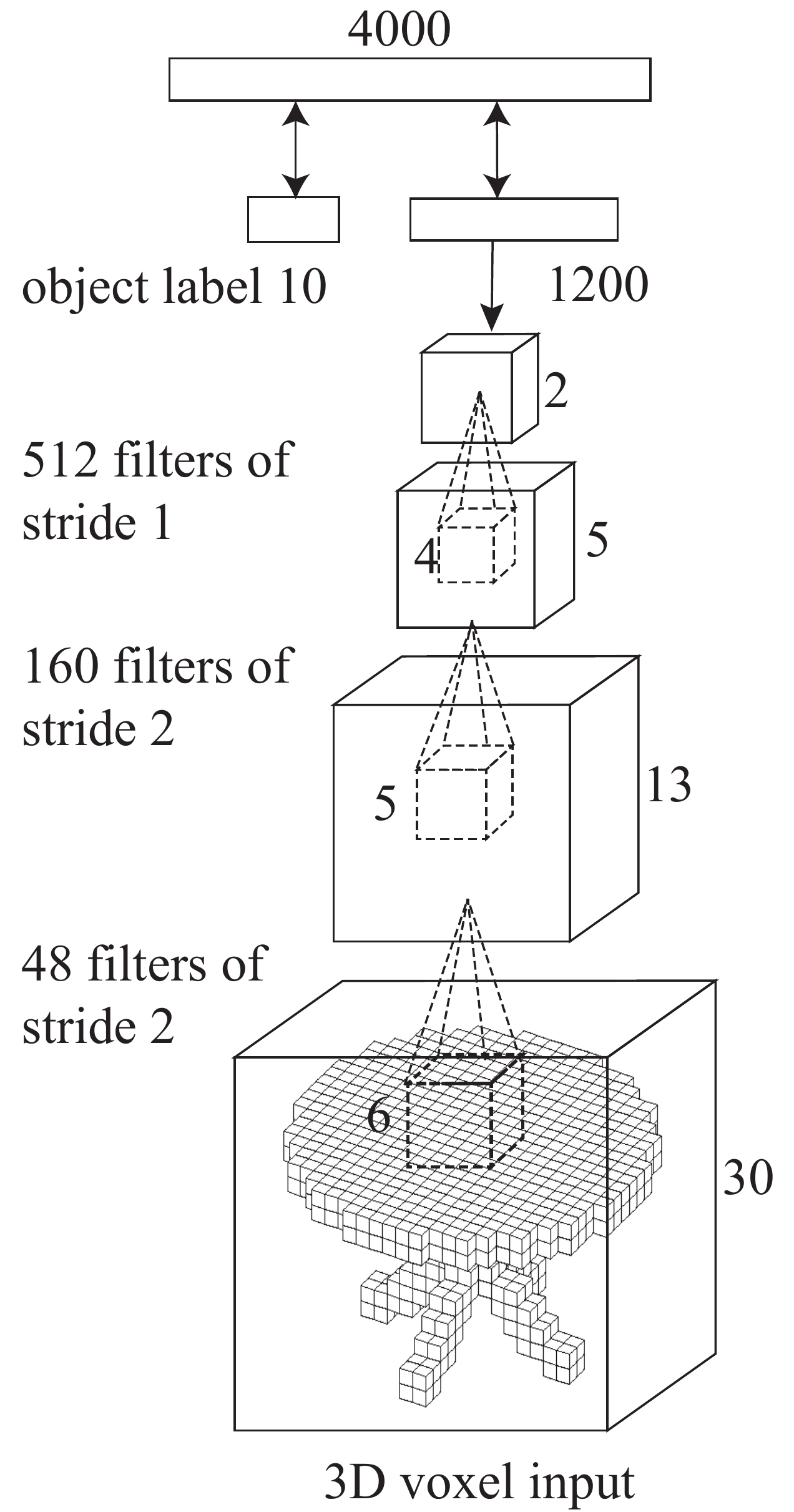}
              \caption{Architecture of our 3D ShapeNets model. For illustration purpose, we only draw one filter for each convolutional layer.}
              \label{fig:architecture}
     \end{subfigure}
     ~
	 \begin{subfigure}{0.71\textwidth}
		\centering
		
		
		\includegraphics[width=0.95\textwidth]{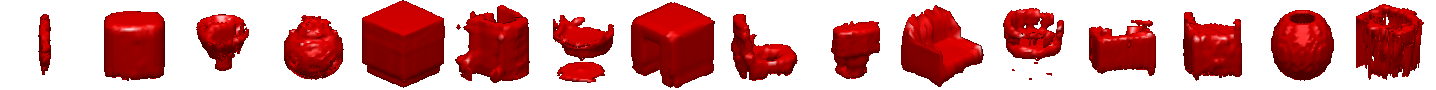}~L5
		
		\vspace{-3mm}\line(1,0){350}
		
		\includegraphics[width=0.95\textwidth]{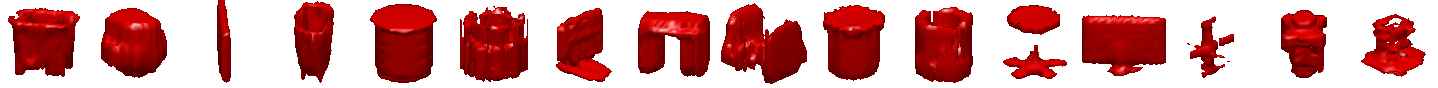}~L4
		
		\vspace{-3mm}\line(1,0){350}
		
		\includegraphics[width=0.95\textwidth]{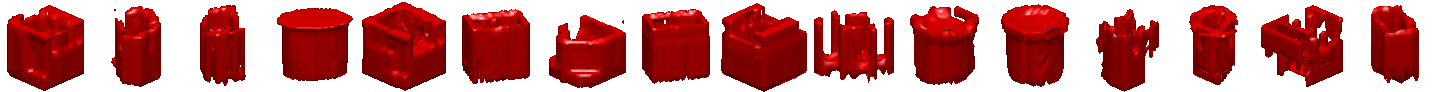}~L3
		
		\vspace{-3mm}\line(1,0){350}
		
		\includegraphics[width=0.95\textwidth]{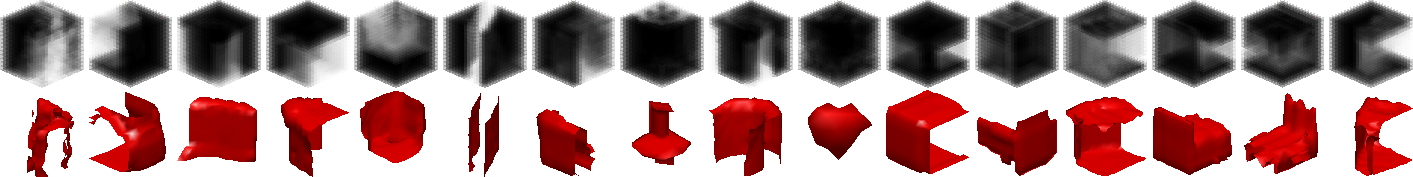}~L2
		
		\vspace{-3mm}\line(1,0){350}
	
	         \includegraphics[width=0.95\textwidth]{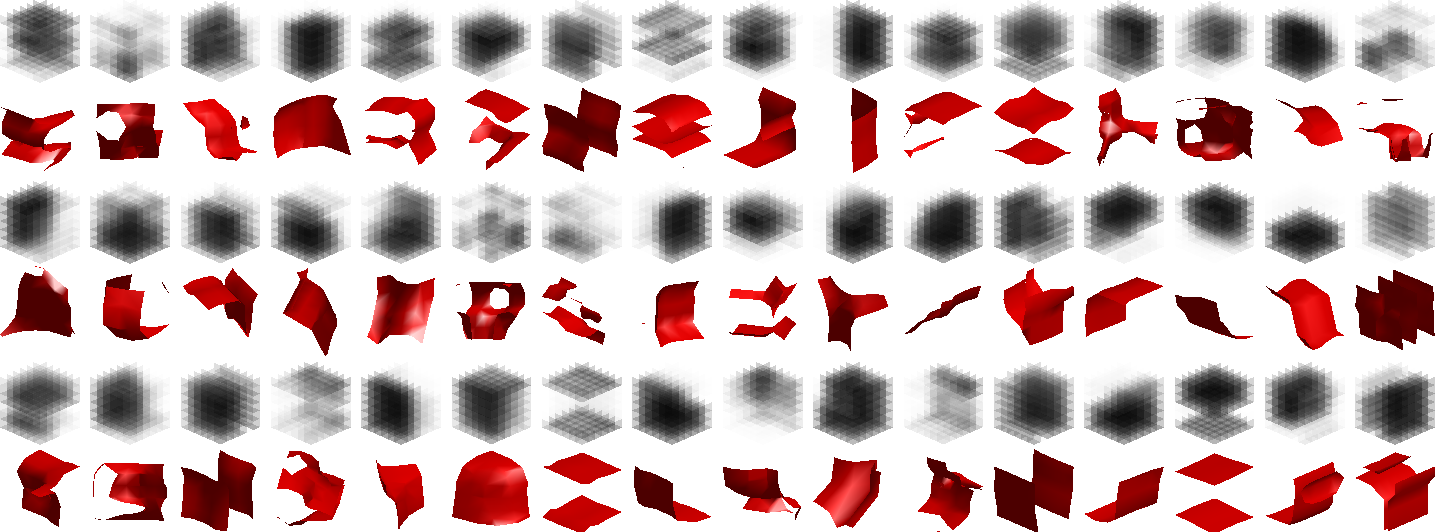}~L1
             \caption{Data-driven visualization: For each neuron, we average the top 100 training examples with highest responses ($>$0.99) and crop the volume inside the receptive field. The averaged result is visualized by transparency in 3D (Gray) and by the average surface obtained from  zero-crossing (Red). 3D ShapeNets are able to capture complex structures in 3D space,
from low-level surfaces and corners at L1, to objects parts at L2 and L3, and whole objects at L4 and above.}
              \label{fig:filters}
     \end{subfigure}

\vspace{-1mm}
\caption{{\bf 3D ShapeNets.} Architecture and filter visualizations from different layers.}
\label{fig:ShapeNets}
\end{figure*}

In this paper, we study generic shape representation for both object category recognition and shape completion. 
While there has been significant progress on shape synthesis~\cite{Sid2011,Sid2012} and recovery~\cite{Shen2012}, they are mostly limited to part-based assembly and 
heavily rely on expensive part annotations.
Instead of hand-coding shapes by parts, 
we desire a data-driven way to learn the complex shape distributions from raw 3D data across object categories and poses,
and automatically discover a hierarchical compositional part representation.
As shown in Figure~\ref{fig:teaser},
this would allow us to infer the full 3D volume from a depth map without 
the knowledge of object category and pose a priori.
Beyond the ability to jointly hallucinate missing structures and predict categories, 
we also desire the ability to compute the potential information gain for recognition with regard to missing parts.
This would allow an active recognition system to choose an optimal subsequent view for observation, when the category recognition from the first view is not sufficiently confident.

%

To this end,
we propose 3D ShapeNets to represent a geometric 3D shape as a probabilistic distribution of binary variables on a 3D voxel grid. Our model 
uses a powerful Convolutional Deep Belief Network (Figure~\ref{fig:ShapeNets}) to learn the complex joint distribution of all 3D voxels in a data-driven manner.
To train this 3D deep learning model, 
we construct ModelNet, a large-scale object dataset of 3D computer graphics CAD models.
We demonstrate the strength of our model at capturing complex object shapes by drawing samples from the model. 
We show that our model can recognize objects in single-view 2.5D depth images and hallucinate the missing parts of depth maps. 
Extensive experiments 
suggest that
our model also generalizes well to real world data from the NYU depth dataset~\cite{NYUdataset}, 
significantly outperforming existing approaches on single-view 2.5D object recognition.
Further it is also effective for next-best-view prediction in view planning for active object recognition~\cite{NBVsurveys}.

\begin{figure*}[t]
\centering
\includegraphics[width = 1\textwidth]{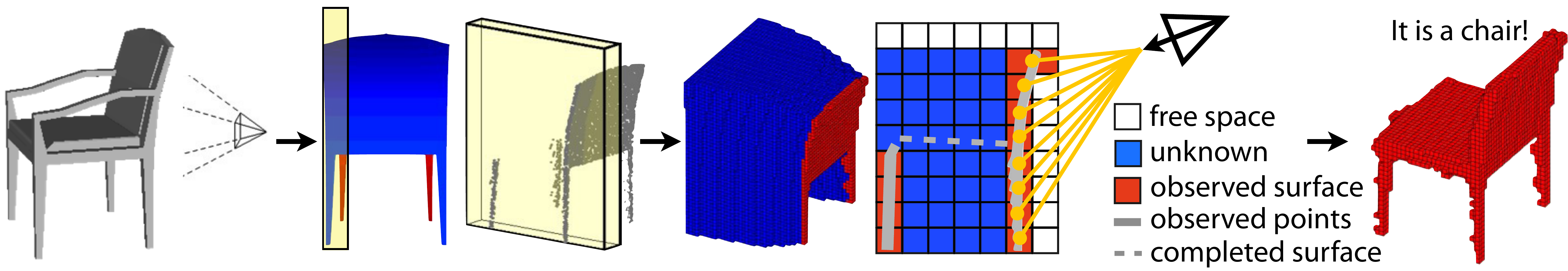}
~~~~~~(1) object~~~~~~~~~~~~~~~~~~~~~~(2) depth \& point cloud~~~~~~~~~~~~~(3) volumetric representation~~~~~~~~~~~~(4) recognition \& completion 

\vspace{-1mm}
\caption{
{\bf View-based 2.5D Object Recognition.}
(1) Illustrates that a depth map is taken from a physical object in the 3D world. 
(2) Shows the depth image captured from the back of the chair. A slice is used for visualization. 
(3) Shows the profile of the slice and different types of voxels.
The surface voxels of the chair $\mathbf{x}_o$ are in red, and the occluded voxels $\mathbf{x}_u$ are in blue.
(4) Shows the recognition and shape completion result, conditioned on the observed free space and surface.}
\label{fig:25Drecognition}
\vspace{-3mm}
\end{figure*}

\section{Related Work}
There has been a large body of insightful research on analyzing 3D CAD model collections. Most of the works~\cite{Sid2011,TomAssembly,Sid2012} use an assembly-based approach to build deformable part-based models. These methods are limited to a specific class of shapes with small variations, with surface correspondence being one of the key problems in such approaches. Since we are interested in shapes across a variety of objects with large variations and part annotation is tedious and expensive, assembly-based modeling can be rather cumbersome.
For surface reconstruction of corrupted scanning input, most related works~\cite{recon2,recon1} are largely based on smooth interpolation or extrapolation. These approaches can only tackle small missing holes or deficiencies. Template-based methods~\cite{Shen2012} are able to deal with large space corruption but are mostly limited by the quality of available templates and often do not provide different semantic interpretations of reconstructions.

The great generative power of deep learning models has allowed researchers to build deep generative models for 2D shapes: most notably the DBN~\cite{DBN} to generate handwritten digits and ShapeBM~\cite{ShapeBM2012} to generate horses, etc. These models are able to effectively capture intra-class variations. We also desire this generative ability for shape reconstruction but we focus on more complex real world object shapes in 3D. 
For 2.5D deep learning, \cite{Socher} and \cite{depthRCNN} build discriminative convolutional neural nets to model images and depth maps. 
Although their algorithms are applied to depth maps, they use depth as
an extra 2D channel instead of modeling full 3D. 
Unlike~\cite{Socher}, our model learns a shape distribution over a voxel grid. To the best of our knowledge, we are the first work to build 3D deep learning models.
To deal with the dimensionality of high resolution voxels, inspired by~\cite{CDBN}\footnote{The model is precisely a convolutional DBM where all the connections are undirected, while ours is a convolutional DBN.}, we apply the same convolution technique in our model.






Unlike static object recognition in a single image, the sensor in active object recognition~\cite{callari2001active} can move to new view points to gain more information about the object. Therefore, the Next-Best-View problem~\cite{NBVsurveys} of doing view planning based on current observation arises. 
Most previous works in active object recognition~\cite{denzler2002information,jia2009active} 
build their view planning strategy using 2D color information. However this multi-view problem is intrinsically 3D in nature.
Atanasov et al,~\cite{atanasov2013hypothesis,atanasov2013nonmyopic} implement the idea in real world robots, but they assume that there is only one object associated with each class reducing their problem to instance-level recognition with no intra-class variance. 
 Similar to~\cite{denzler2002information},  we use mutual information to decide the NBV. However, we consider this problem at the precise voxel level allowing us to infer how voxels in a 3D region would contribute to the reduction of recognition uncertainty.

\section{3D ShapeNets} 

To study 3D shape representation,
we propose to represent a geometric 3D shape as a probability distribution of binary variables on a 3D voxel grid.
Each 3D mesh is represented as a binary tensor: 1 indicates the voxel is inside the mesh surface, and 0 indicates the voxel is outside the mesh (i.e., it is empty space).
The grid size in our experiments is $30\times30\times30$. 

To represent the probability distribution of these binary variables for 3D shapes,
we design a Convolutional Deep Belief Network (CDBN).
Deep Belief Networks (DBN) \cite{DBN} are a powerful class of probabilistic models often used to model the joint probabilistic distribution over pixels and labels in 2D images. 
Here, we adapt the model from 2D pixel data to 3D voxel data, 
which imposes some unique challenges. A 3D voxel volume with reasonable resolution (say $30\times30\times30$) would have the same dimensions as a high-resolution image ($165\times 165$). A fully connected DBN on such an image would result in a huge number of parameters making the model intractable to train effectively. 
Therefore, we propose to use convolution to reduce model parameters by weight sharing.
However, different from typical convolutional deep learning models (e.g. \cite{CDBN}),
we do not use any form of pooling in the hidden layers --
while pooling may enhance the invariance properties for recognition, in our case, it would also lead to greater uncertainty for shape reconstruction.

The energy, $E$, of a convolutional layer in our model can be computed as:
\begin{equation}
E(\mathbf{v},\mathbf{h}) = -\sum_f \sum_j \left(h_j^f \left( W^f *  v\right)_j + c^f h_j^f  \right) - \sum_l b_lv_l
\end{equation}
where $v_l$ denotes each visible unit, $h_j^f$ denotes each hidden unit in a feature channel $f$, and $W^f$ denotes the convolutional filter. The ``$\ast$'' sign represents the convolution operation. In this energy definition, each visible unit $v_l$ is associated with a unique bias term $b_l$ to facilitate reconstruction, and all hidden units $\{h_j^f\}$ in the same convolution channel share the same bias term $c^f$. Similar to~\cite{DCNN}, we also allow for a convolution stride.

\begin{figure}[t]
\includegraphics[width=1\linewidth]{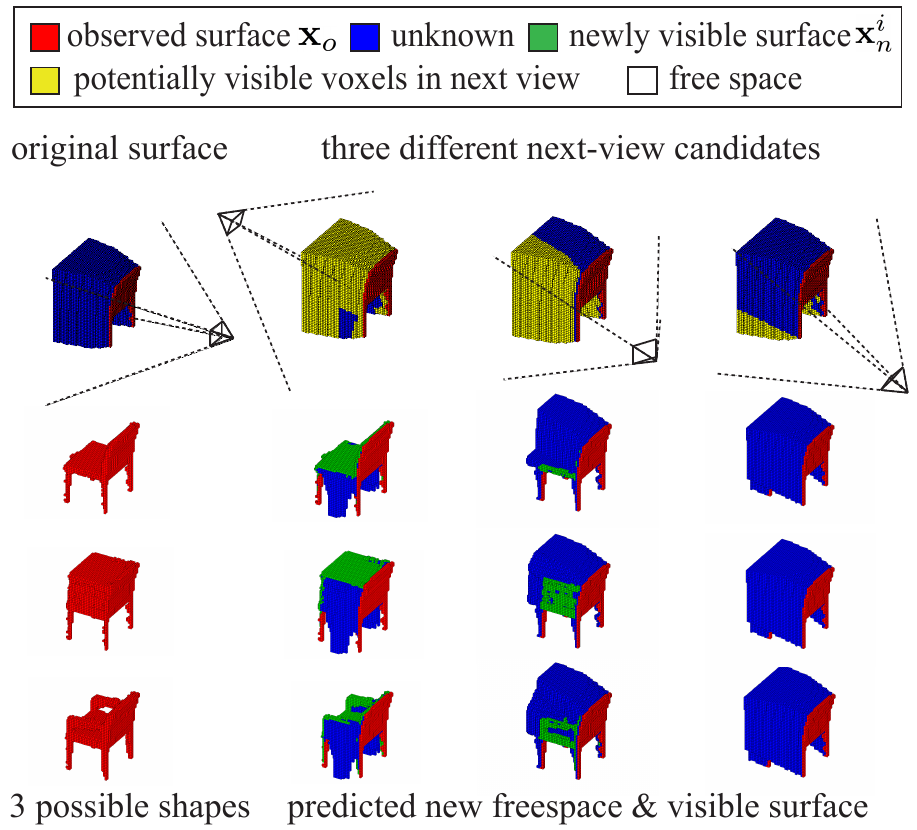}

\vspace{-2mm}
\caption{{\bf Next-Best-View Prediction.} 
[Row 1, Col 1]: the observed (red) and unknown (blue) voxels from a single view.
[Row 2-4, Col 1]: three possible completion samples generated by conditioning on $(\mathbf{x}_o,\mathbf{x}_u)$. 
[Row 1, Col 2-4]: three possible camera positions $\mathbf{V}^i$, front top, left-sided, tilted bottom, front, top. 
[Row 2-4, Col 2-4]: predict the new visibility pattern of the object given the possible shape and camera position $\mathbf{V}^i$.}
\label{fig:nextview}

\vspace{-3mm}
\end{figure}

A 3D shape is represented as a $24\times24\times24$ voxel grid with 3 extra cells of padding in both directions to reduce the convolution border artifacts. The labels are presented as standard one of $K$ softmax variables.
The final architecture of our model is illustrated in Figure~\ref{fig:ShapeNets}(a). The first layer has 48 filters of size 6 and stride 2; the second layer has 160 filters of size 5 and stride 2 
(i.e., each filter has $48\hspace{-1mm}\times\hspace{-1mm}5\hspace{-1mm}\times\hspace{-1mm}5\hspace{-1mm}\times\hspace{-1mm}5$ parameters); the third layer has 512 filters of size 4; each convolution filter is connected to all the feature channels in the previous layer; the fourth layer is a standard fully connected RBM with 1200 hidden units; and the fifth and final layer with 4000 hidden units takes as input a combination of multinomial label variables and Bernoulli feature variables. The top layer forms an associative memory DBN as indicated by the bi-directional arrows, while all the other layer connections are directed top-down. 

We first pre-train the model in a layer-wise fashion followed by a generative fine-tuning procedure. During pre-training, the first four layers are trained using standard Contrastive Divergence~\cite{CD}, while the top layer is trained more carefully using Fast Persistent Contrastive Divergence (FPCD)~\cite{FPCD}. Once the lower layer is learned, the weights are fixed and the hidden activations are fed into the next layer as input. Our fine-tuning procedure is similar to wake sleep algorithm~\cite{DBN} except that we keep the weights tied. In the wake phase, we propagate the data bottom-up and use the activations to collect the positive learning signal. In the sleep phase, we maintain a persistent chain on the topmost layer and propagate the data top-down to collect the negative learning signal. This fine-tuning procedure mimics the recognition and generation behavior of the model and works well in practice. 
We visualize some of the learned filters in Figure~\ref{fig:ShapeNets}(b).

During pre-training of the first layer, we collect learning signal only in receptive fields which are non-empty. Because of the nature of the data, empty spaces occupy a large proportion of the whole volume, which have no information for the RBM and would distract the learning. Our experiment shows that ignoring those learning signals during gradient computation results in our model learning more meaningful filters. 
In addition, for the first layer, we also add sparsity regularization to restrict the mean activation of the hidden units to be a small constant (following the method of~\cite{Sparsity}).
During pre-training of the topmost RBM where the joint distribution of labels and high-level abstractions are learned, we duplicate the label units 10 times to increase their significance.


\section{2.5D Recognition and Reconstruction}

\subsection{View-based Sampling}

\begin{figure*}[t]
\vspace{-2mm}

\includegraphics[width=1\linewidth]{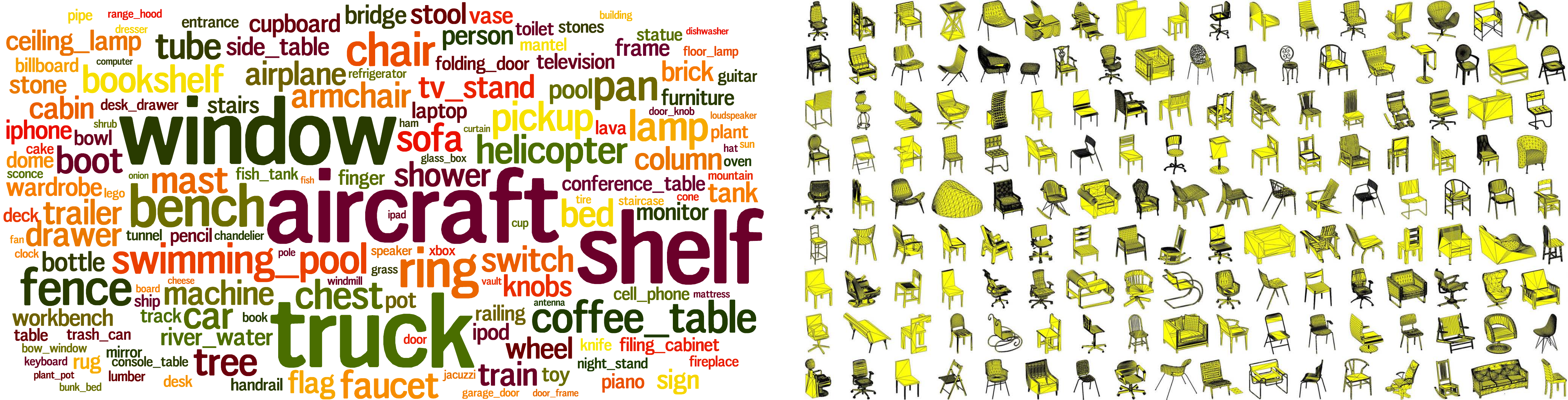}

\vspace{-2mm}
\caption{{\bf ModelNet Dataset.} Left: word cloud visualization of the ModelNet dataset based on the number of
3D models in each category. Larger font size indicates more instances in the category. Right: Examples of 3D chair models.}
\label{fig:modelnet}
\vspace{-3mm}
\end{figure*}

After training the CDBN, the model learns the joint distribution $p(\mathbf{x},y)$ of voxel data $\mathbf{x}$ and object category label $y \in \{ 1,\cdots,K\}$. 
Although the model is trained on complete 3D shapes,
it is able to recognize objects in single-view 2.5D depth maps (e.g., from RGB-D sensors).
As shown in Figure \ref{fig:25Drecognition},
the 2.5D depth map is first converted into a volumetric representation where we categorize each voxel as free space, surface or occluded,
depending on whether it is in front of, on, or behind the visible surface (i.e., the depth value) from the depth map.
The free space and surface voxels are considered to be observed, and the occluded voxels are regarded as missing data.
The test data is represented by $\mathbf{x}=(\mathbf{x}_o,\mathbf{x}_u)$, where $\mathbf{x}_o$ refers to the observed free space and surface voxels,
while $\mathbf{x}_u$ refers to the unknown voxels.
Recognizing the object category involves estimating $p(y|\mathbf{x}_o)$. 
We approximate the posterior distribution $p(y|\mathbf{x}_o)$ by Gibbs sampling. The sampling procedure is as follows. We first initialize $\mathbf{x}_u$ to a random value and propagate the data $\mathbf{x} = (\mathbf{x}_o,\mathbf{x}_u)$ bottom up to sample for a label $y$ from $p(y|\mathbf{x}_o, \mathbf{x}_u)$. Then the high level signal is propagated down to sample for voxels $\mathbf{x}$. We clamp the observed voxels $\mathbf{x}_o$ in this sample $\mathbf{x}$ and do another bottom up pass. 50 iterations of up-down sampling are sufficient to get a shape completion $\mathbf{x}$, and its corresponding label $y$. 
The above procedure is run in parallel for a large number of particles resulting in a variety of completion results corresponding to potentially different classes. The final category label corresponds to the most frequently sampled class.

\subsection{Next-Best-View Prediction}

Object recognition from a single-view can sometimes be challenging, both for humans and computers. 
However, if an observer is allowed to view the object from another view point when recognition fails from the first view point, we may be able to significantly reduce the recognition uncertainty. 
Given the current view, our model is able to predict which next view would be optimal for discriminating the object category.

The inputs to our next-best-view system are observed voxels $\mathbf{x}_o$ of an unknown object captured by a depth camera from a single view, and  a finite list of next-view candidates $\{\mathbf{V}^i\}$ representing the camera rotation and translation in 3D.
An algorithm chooses the next-view from the list 
that has the highest potential to reduce the recognition uncertainty. 
Note that during this view planning process, we do not observe any new data, and hence there is no improvement on the confidence of $p(y|\mathbf{x}_o=x_o)$. 

The original recognition uncertainty, $H$, is given by the entropy of $y$ conditioned on the observed $\mathbf{x}_o$:
\begin{equation}
\begin{split}
H &= H\left(p(y|\mathbf{x}_o=x_o)\right) \\
&= -\sum_{k=1}^{K} p(y=k|\mathbf{x}_o=x_o) \textrm{log }p(y=k|\mathbf{x}_o=x_o)
\end{split}
\end{equation}
where the conditional probability $p(y|\mathbf{x}_o=x_o)$ can be approximated as before by sampling from $p(y,\mathbf{x}_u|\mathbf{x}_o=x_o)$ and marginalizing $\mathbf{x}_u$.

When the camera is moved to another view $\mathbf{V}^i$,
some of the previously unobserved voxels $\mathbf{x}_u$ may become observed based on its actual shape. 
Different views $\mathbf{V}^i$ will result in different visibility of these unobserved voxels $\mathbf{x}_u$.
A view with the potential to see distinctive parts of objects (e.g. arms of chairs) 
may be a better next view.
However, since the actual shape is partially unknown\footnote{If the 3D shape is fully observed, adding more views will not help to reduce the recognition uncertainty in any algorithm purely based on 3D shapes,
including our 3D ShapeNets.},
we will hallucinate that region from our model.
As shown in Figure~\ref{fig:nextview},
conditioning on $\mathbf{x}_o=x_o$,
we can sample many shapes to generate hypotheses of the actual shape, 
and then render each hypothesis to obtain the depth maps observed from different views, $\mathbf{V}^i$.
In this way, we can simulate the new depth maps for different views on different samples
and compute the potential reduction in recognition uncertainty.

Mathematically, 
let $\mathbf{x}_n^i = \textrm{Render}(\mathbf{x}_u, \mathbf{x}_o, \mathbf{V}^i ) \setminus \mathbf{x}_o$
 denote the {\bf new} observed voxels 
(both free space and surface) in the next view $\mathbf{V}^i$.
We have $\mathbf{x}_n^i \subseteq \mathbf{x}_u$, and they are unknown variables that will be marginalized in the following equation. 
Then the potential recognition uncertainty for $\mathbf{V}^i$ is measured by this conditional entropy,
\begin{equation}
\begin{split}
H_{i} &= H\left(p(y|\mathbf{x}_n^i,\mathbf{x}_o=x_o)\right) \\
&= \sum_{\mathbf{x}_n^i} p(\mathbf{x}_n^i|\mathbf{x}_o=x_o)H(y|\mathbf{x}_n^i,\mathbf{x}_o=x_o).
\end{split}
\end{equation}
The above conditional entropy could be calculated by first sampling enough $\mathbf{x}_u$ from $p(\mathbf{x}_u|\mathbf{x}_o=x_o)$,
doing the 3D rendering to obtain 2.5D depth map in order to get $\mathbf{x}_n^i$ from $\mathbf{x}_u$, 
and then taking each $\mathbf{x}_n^i$ to calculate $H(y|\mathbf{x}_n^i=x_n^i,\mathbf{x}_o=x_o)$ as before. 


According to information theory, the reduction of entropy
$H - H_i = I(y;\mathbf{x}_n^i|\mathbf{x}_o=x_o) \geq 0$ is the mutual information between $y$ and $\mathbf{x}_n^i$ conditioned on $\mathbf{x}_o$. This meets our intuition that observing more data will always potentially reduce the uncertainty. 
With this definition, 
our view planning algorithm is to simply choose the view that maximizes this mutual information,
\begin{equation}
\mathbf{V}^* =   {\arg\max}_{\mathbf{V}^i} I(y;\mathbf{x}_n^i|\mathbf{x}_o=x_o).
\end{equation} 

Our view planning scheme can naturally be extended to a sequence of view planning steps. After deciding the best candidate to move for the first frame, we physically move the camera there and capture the other object surface from that view. The object surfaces from all previous views are merged together as our new observation $\mathbf{x}_o$, allowing us to run our view planning scheme again.

\begin{figure}[t]
\vspace{-2mm}

\footnotesize
\centering
\includegraphics[width=1\linewidth]{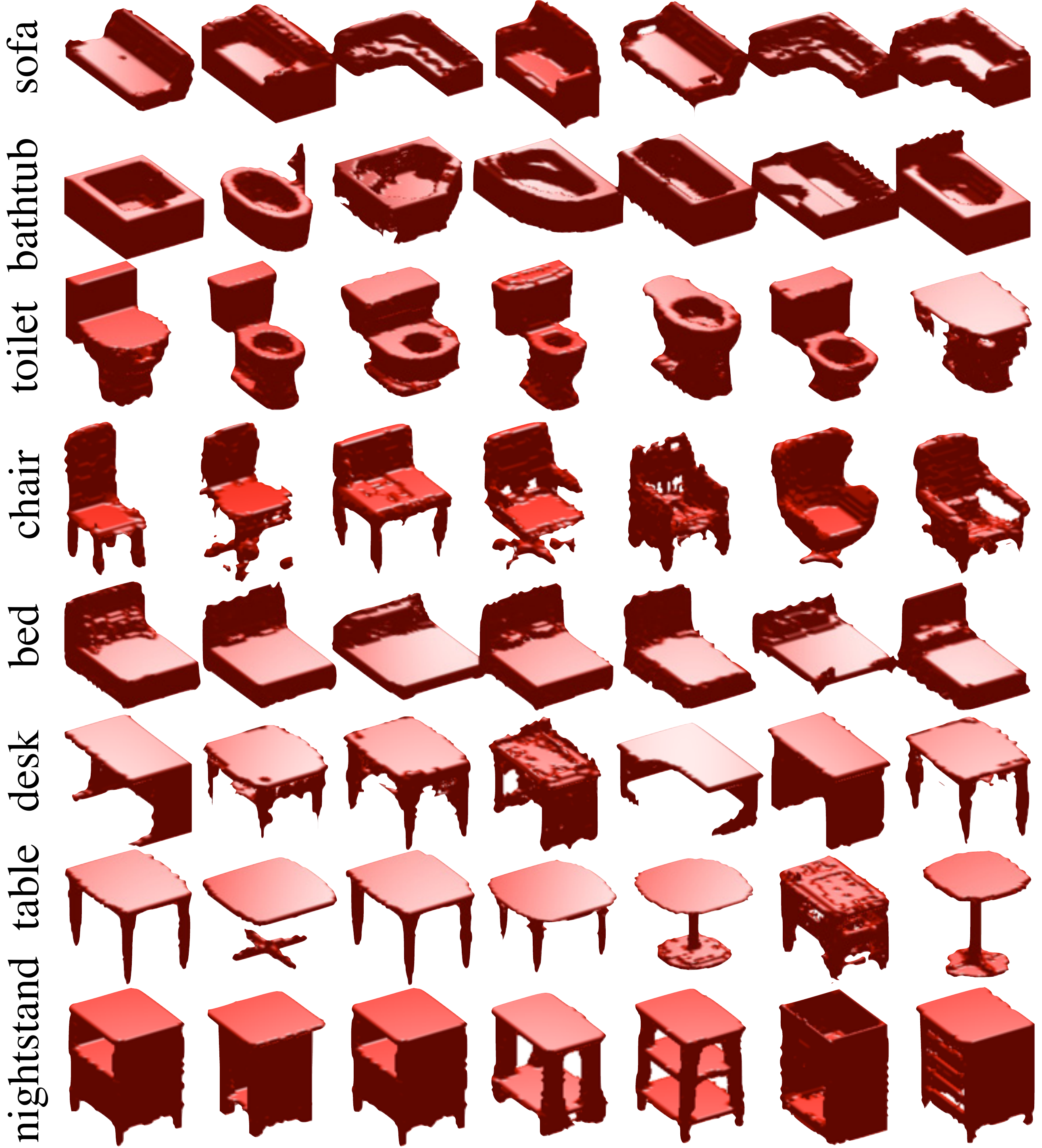}

\vspace{-1mm}
\caption{{\bf Shape Sampling.} Example shapes generated by sampling our 3D ShapeNets for some categories.} 
\label{fig:samples}
\end{figure}

\section{ModelNet: A Large-scale 3D CAD Dataset}
Training a deep 3D shape representation that captures intra-class variance requires a large collection of 3D shapes. 
Previous CAD datasets (e.g.,~\cite{Psb}) are limited both in the variety of categories and the number of examples per category. 
Therefore, we construct ModelNet, a large-scale 3D CAD model dataset.

To construct ModelNet, we downloaded 3D CAD models from 3D Warehouse, and Yobi3D search engine indexing 261 CAD model websites. 
We query common object categories from the SUN database \cite{SUNDB}
that contain no less than 20 object instances per category, removing
those with too few search results, resulting in a total of 660 categories. We also include models from the Princeton Shape Benchmark \cite{Psb}.
After downloading, we remove mis-categorized models using Amazon
Mechanical Turk. Turkers are shown a sequence of thumbnails of the
models and answer ``Yes'' or ``No'' as to whether the category label
matches the model. The authors then manually checked each 3D model and
removed irrelevant objects from each CAD model (e.g, floor, thumbnail
image, person standing next to the object, etc) so that each mesh
model contains only one object belonging to the labeled category. We
also discarded unrealistic (overly simplified models or those only containing images of the object) and duplicate models.
Compared to~\cite{Psb}, which consists of 6670 models in 161 categories, our new dataset is 22 times larger containing 151,128 3D CAD models belonging to 660 unique object categories. Examples of major categories and dataset statistics are shown in Figure \ref{fig:modelnet}.


\section{Experiments}
We choose 40 common object categories from ModelNet with 100 unique CAD models per category. We then augment the data by rotating each model every 30 degrees along the gravity direction (i.e., 12 poses per model) resulting in models in arbitrary poses. Pre-training and fine-tuning each took about two days on a desktop with one Intel XEON E5-2690 CPU and one NVIDIA K40c GPU.
Figure \ref{fig:samples} shows some shapes sampled from our trained model.



\subsection{3D Shape Classification and Retrieval}
\label{sec:exp:classification}

Deep learning has been widely used as a feature extraction technique. 
Here, we are also interested in how well the features learned from 3D ShapeNets compare with other state-of-the-art 3D mesh features. We discriminatively fine-tune 3D ShapeNets by replacing the top layer with class labels and use the 5th layer as features. For comparison, we choose Light Field descriptor~\cite{LFDfeature} (LFD, 4,700 dimensions) and Spherical Harmonic descriptor~\cite{SHPfeature} (SPH, 544 dimensions), which performed best among all descriptors~\cite{Psb}. 

\begin{table}[t]
\centering
\begin{tabular}{c|c|c|c}

\Xhline{2\arrayrulewidth}
10 classes & SPH~\cite{SHPfeature} & LFD~\cite{LFDfeature} & Ours\tabularnewline
\Xhline{2\arrayrulewidth}
classification & 79.79 \% & 79.87 \% & {\bf 83.54}\%
\tabularnewline
retrieval AUC & 45.97\% & 51.70\% & {\bf 69.28}\%
\tabularnewline
retrieval MAP & 44.05\% & 49.82\% & {\bf 68.26}\%
\tabularnewline

\Xhline{2\arrayrulewidth}
40 classes & SPH~\cite{SHPfeature} & LFD~\cite{LFDfeature} & Ours\tabularnewline
\Xhline{2\arrayrulewidth}
classification & 68.23\% & 75.47\% & {\bf 77.32}\%
\tabularnewline
retrieval AUC & 34.47\% & 42.04\% & {\bf 49.94}\%
\tabularnewline
retrieval MAP & 33.26\% & 40.91\% & {\bf 49.23}\%
\tabularnewline

\Xhline{2\arrayrulewidth}
\end{tabular}

\vspace{-2mm}
\caption{{\bf Shape Classification and Retrieval Results.} }
\vspace{-2mm}
\label{table:cls}
\end{table}

We conduct 3D classification and retrieval experiments to evaluate our features. Of the 48,000 CAD models (with rotation enlargement), 38,400 are used for training and 9,600 for testing. We also report a smaller scale result on a 10-category subset (corresponding to NYU RGB-D dataset~\cite{NYUdataset}) of the 40-category data. For classification, we train a linear SVM to classify meshes using each of the features mentioned above, and use average category accuracy to evaluate the performance. 


For retrieval, we use $L2$ distance to measure the similarity of the shapes between each pair of testing samples. Given a query from the test set, a ranked list of the remaining test data is returned according to the similarity measure\footnote{For our feature and SPH we use the $L2$ norm, and for LFD we use the distance measure from~\cite{LFDfeature}.}. We evaluate retrieval algorithms using two metrics: (1) mean area under precision-recall curve (AUC) for all the testing queries\footnote{We interpolate each precision-recall curve.}; (2) mean average precision (MAP) where AP is defined as the average precision each time a positive sample is returned. 

\begin{figure}[t]
\centering
\includegraphics[width=0.43\linewidth]{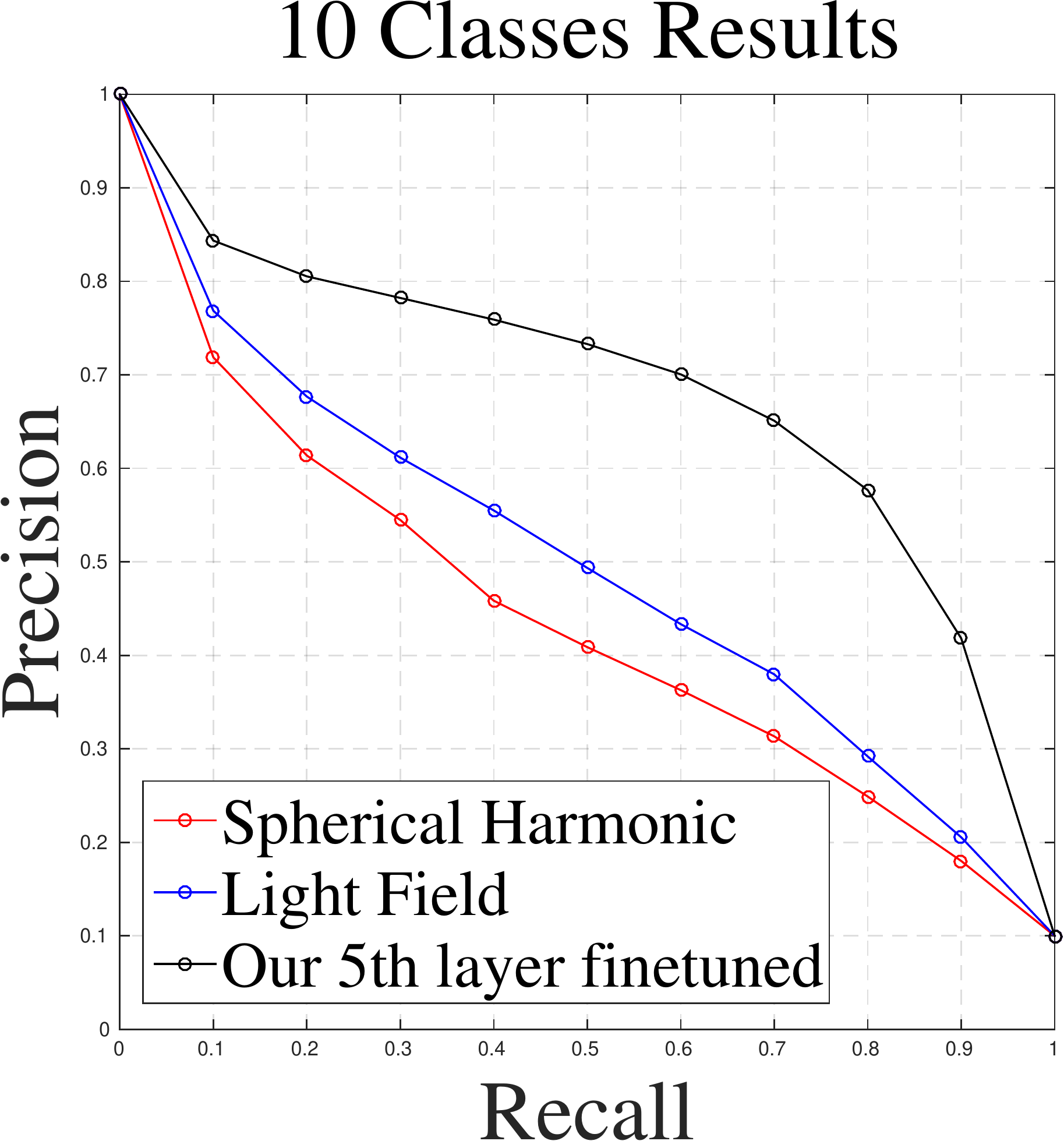} \quad
\includegraphics[width=0.43\linewidth]{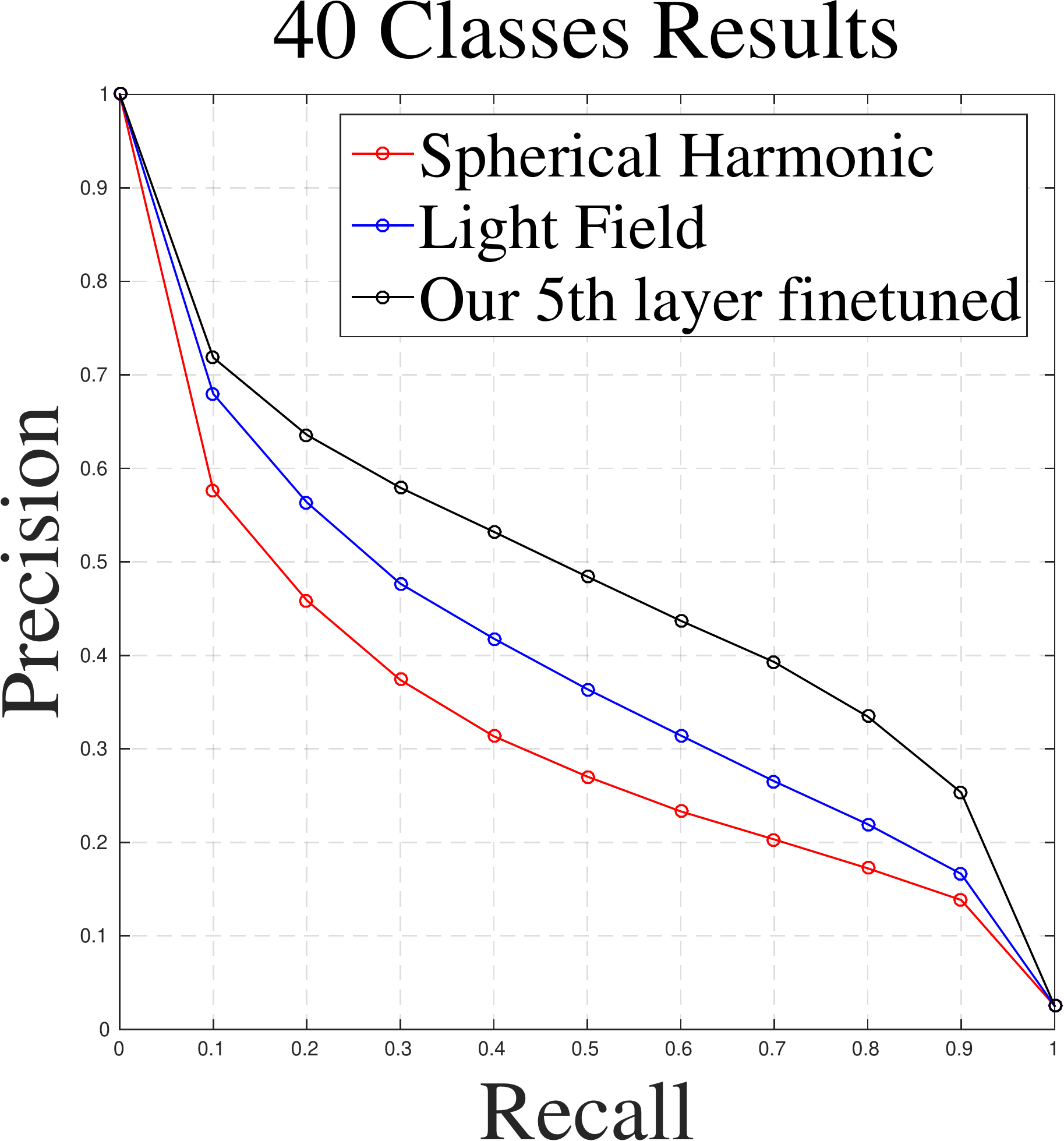}


\vspace{-2mm}
\caption{{\bf 3D Mesh Retrieval.} Precision-recall curves at standard recall levels.}
\label{fig:feature}
\end{figure}

We summarize the results in Table~\ref{table:cls} and Figure~\ref{fig:feature}. 
Since both of the baseline mesh features (LFD and SPH) are rotation invariant, from the performance we have achieved, we believe 3D ShapeNets must have learned this invariance during feature learning. Despite using a significantly lower resolution mesh as compared to the baseline descriptors, 3D ShapeNets outperforms them by a large margin. 
This demonstrates that our 3D deep learning model can learn better features from 3D data automatically.

\subsection{View-based 2.5D Recognition}


To evaluate 3D ShapeNets for 2.5D depth-based object recognition task, 
we set up an experiment on the NYU RGB-D dataset with Kinect depth maps \cite{NYUdataset}.
We select 10 object categories from ModelNet that overlap with the NYU dataset. This results in 4,899 unique CAD models for training 3D ShapeNets. 

\begin{figure*}[t]
\includegraphics[width=1\linewidth]{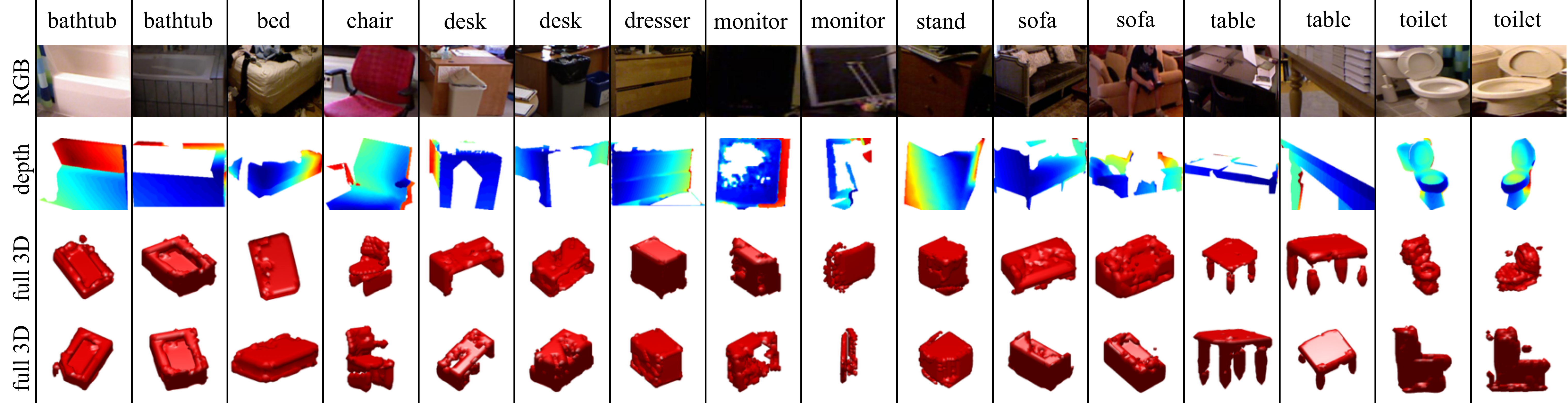}
\vspace{-5mm}
\caption{{\bf Successful Cases of Recognition and Reconstruction
on NYU dataset} \cite{NYUdataset}.
In each example, we show the RGB color crop, the segmented depth map, and the shape reconstruction from two view points.}
\label{fig:NYUvis}
\end{figure*}

We create each testing example by cropping the 3D point cloud from the 3D bounding boxes. 
The segmentation mask is used to remove outlier depth in the bounding box.
Then we directly apply our model trained on CAD models to the NYU dataset. This is absolutely non-trivial because the statistics of real world depth are significantly different from the synthetic CAD models used for training. In Figure \ref{fig:NYUvis}, we visualize the successful recognitions and reconstructions. Note that 3D ShapeNets is even able to partially reconstruct the ``monitor'' despite the bad scanning caused by the reflection problem. To further boost recognition performance, we discriminatively fine-tune our model on the NYU dataset
using back propagation. 
By simply assigning invisible voxels as 0 (i.e. considering occluded voxels as free space and only representing the shape as the voxels on the 3D surface) and rotating training examples every 30 degrees, fine-tuning works reasonably well in practice.


As a baseline approach, we use $k$-nearest-neighbor matching in our low resolution voxel space. Testing depth maps are converted to voxel representation and compared with each of the training samples.
As a more sophisticated high resolution baseline, we match the testing point cloud to each of our 3D mesh models using Iterated Closest Point method \cite{ICP} and use the top 10 matches to vote for the labels. 
We also compare our result with \cite{Socher} which is the state-of-the-art deep learning model applied to RGB-D data. 
To train and test their model, 2D bounding boxes are obtained by projecting the 3D bounding box to the image plane, and object segmentations are also used to extract features. 
1,390 instances are used to train the algorithm of~\cite{Socher} and perform our discriminative fine-tuning, while the remaining 495 instances
are used for testing all five methods. 
Table \ref{table:ap} summarizes the recognition results. 
Using only depth without color,
our fine-tuned 3D ShapeNets outperforms all other approaches with or without color by a significant margin.

\begin{figure}[t]
\vspace{-2mm}

{\small
~~~~Input~~~~~~~~GT~~~~~~~~~3D ShapeNets Completion Result~~~~~~NN
}

\vspace{-0.5mm}
\centering
\includegraphics[width=1\linewidth]{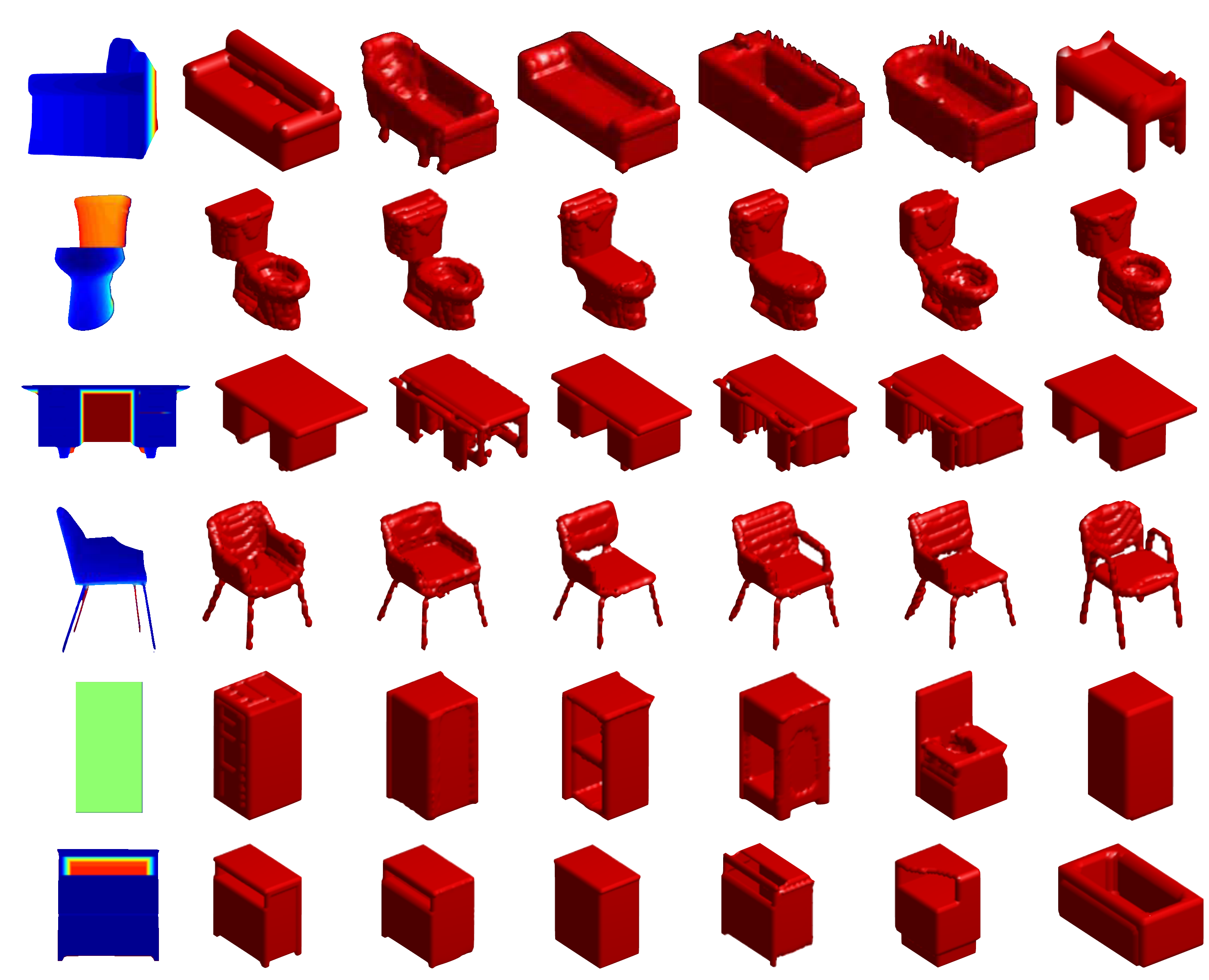}

\vspace{-3mm}
\caption{{\bf Shape Completion.} From left to right: input depth map from a single view, ground truth shape, shape completion result (4 cols), nearest neighbor result (1 col).}
\label{fig:shapecompletion}
\vspace{-3mm}
\end{figure}

\begin{table*}[t]
\centering
\setlength{\tabcolsep}{7.8pt}
{
\centering
\footnotesize
\begin{tabular}{c|c|c|c|c|c|c|c|c|c|c|c}
\Xhline{2\arrayrulewidth}
 & bathtub & bed & chair & desk & dresser & monitor & \hspace{-1mm}nightstand\hspace{-1mm} & sofa & table & toilet & all\tabularnewline
\Xhline{3\arrayrulewidth}
\cite{Socher} Depth & 0.000 & 0.729 & 0.806 & 0.100 & 0.466 & 0.222 & 0.343 & 0.481 & 0.415 & 0.200 & 0.376\tabularnewline
\hline 
NN & 0.429  & 0.446 & 0.395 & 0.176 & 0.467 & 0.333 & 0.188 & 0.458 & 0.455 & 0.400 & 0.374\tabularnewline
\hline 
ICP & 0.571  & 0.608 & 0.194 & {\bf 0.375} & {\bf 0.733} & 0.389 & 0.438 & 0.349 & 0.052 & {\bf 1.000} & 0.471\tabularnewline
\hline 
3D ShapeNets & 0.142  & 0.500 & 0.685 &0.100  & 0.366& {\bf 0.500} &  {\bf 0.719} &  0.277& 0.377 & 0.700 & 0.437 \tabularnewline
\hline 
3D ShapeNets fine-tuned & {\bf 0.857}  & 0.703 & {\bf 0.919} & 0.300  & 0.500 & {\bf 0.500} &  0.625 &  {\bf 0.735} & 0.247 & 0.400 & {\bf 0.579} \tabularnewline
\Xhline{3\arrayrulewidth}
\cite{Socher} RGB  & 0.142 & {\bf 0.743} & 0.766 & 0.150 & 0.266 & 0.166 & 0.218 & 0.313 & 0.376 & 0.200 & 0.334\tabularnewline
\hline 
\cite{Socher}  RGBD & 0.000 & {\bf 0.743} & 0.693 & 0.175 & 0.466 & 0.388 & 0.468 & 0.602 & {\bf 0.441} & 0.500 & 0.448\tabularnewline
\Xhline{2\arrayrulewidth}
\end{tabular}
}

\vspace*{-2mm}
\caption{{\bf Accuracy for View-based 2.5D Recognition on NYU dataset} \cite{NYUdataset}.
The first five rows are algorithms that use only depth information. 
The last two rows are algorithms that also use color information.
Our 3D ShapeNets as a generative model performs reasonably well as compared to the other methods. After discriminative fine-tuning, our method achieves the best performance by a large margin of over 10\%.
}
\label{table:ap}


\vspace*{5mm}
\centering
\setlength{\tabcolsep}{9.4pt}
{
\centering
\footnotesize
\begin{tabular}{c|c|c|c|c|c|c|c|c|c|c|c}
\Xhline{2\arrayrulewidth}
 & bathtub & bed & chair & desk & dresser & monitor & \hspace{-1mm}nightstand\hspace{-1mm}  & \hspace{1mm}sofa\hspace{1mm} & table & toilet & all\tabularnewline
\Xhline{2\arrayrulewidth}
Ours & 0.80 & {\bf 1.00} & {\bf 0.85} & 0.50 & {\bf }0.45 & 0.85 & {\bf 0.75}   & {\bf 0.85} & 0.95 & {\bf 1.00} & {\bf 0.80}\tabularnewline
\hline 
Max Visibility & {\bf 0.85}  &  0.85  & {\bf  0.85} & 0.50 & {\bf 0.45} & {\bf 0.85} & {\bf 0.75} & {\bf 0.85} & 0.90 &  0.95  & 0.78\tabularnewline
\hline 
Furthest Away & 0.65 &  0.85  & 0.75 & {\bf 0.55} & 0.25 & 0.85  &  0.65 & 0.50 & {\bf 1.00} &  0.85 & 0.69\tabularnewline
\hline 
Random Selection & 0.60  &  0.80  & 0.75 & 0.50 & {\bf 0.45} & {\bf 0.90} & 0.70 & 0.65 & 0.90  & 0.90 & 0.72\tabularnewline
\Xhline{2\arrayrulewidth}
\end{tabular}
}
\vspace*{-2mm}
\caption{{\bf Comparison of Different Next-Best-View Selections Based on Recognition Accuracy from Two Views.}
Based on an algorithm's choice, we obtain the actual depth map for the next view and recognize 
the object using those two views in our 3D ShapeNets representation.
}
\label{table:nbv}
\end{table*}

\subsection{Next-Best-View Prediction}



%

For our view planning strategy, computation of the term $p(\mathbf{x}_n^i|\mathbf{x}_o=x_o)$ is critical. When the observation $\mathbf{x}_o$ is ambiguous, samples drawn from $p(\mathbf{x}_n^i|\mathbf{x}_o=x_o)$ should come from a variety of different categories. When the observation is rich, samples should be limited to very few categories. Since $\mathbf{x}_n^i$ is the surface of the completions, we could just test the shape completion performance $p(\mathbf{x}_u|\mathbf{x}_o=x_o)$. In Figure~\ref{fig:shapecompletion}, our results give reasonable shapes across different categories. 
We also match the nearest neighbor in the training set to show that our algorithm is not just memorizing the shape and it can generalize well.

To evaluate our view planning strategy, we use CAD models from the test set to create synthetic renderings of depth maps. 
We evaluate the accuracy by running our 3D ShapeNets model on the integration depth maps of both the first view and the selected second view.
A good view-planning strategy should result in a better recognition accuracy.
Note that next-best-view selection is always coupled with the recognition algorithm.
We prepare three baseline methods for comparison : (1) random selection among the candidate views; 
(2) choose the view with the highest new visibility (yellow voxels, NBV for reconstruction); 
(3) choose the view which is farthest away from the previous view (based on camera center distance). 
In our experiment, we generate 8 view candidates randomly distributed on the sphere of the object, 
pointing to the region near the object center and, we randomly choose 200 test examples (20 per category) from our testing set. 
Table \ref{table:nbv} reports the recognition accuracy of different view planning strategies with the same recognition 3D ShapeNets.
We observe that our entropy based method outperforms all other strategies. 

\section{Conclusion}

To study 3D shape representation for objects,
we propose a convolutional deep belief network to
represent a geometric 3D shape as a probability distribution of binary variables on a 3D voxel grid.
Our model can jointly recognize and reconstruct objects from a single-view 2.5D depth map (e.g. from popular RGB-D sensors).
To train this 3D deep learning model, 
we construct ModelNet, a large-scale 3D CAD model dataset.
Our model significantly outperforms existing approaches on a variety
of recognition tasks, and it is also a promising approach for next-best-view planning.
All source code and data set are available at our project website.

\paragraph{Acknowledgment.}
This work is supported by gift funds from Intel Corporation and Project X grant to the Princeton Vision Group,
and a hardware donation from NVIDIA Corporation.
Z.W. is also partially supported by Hong Kong RGC Fellowship.
We thank 
Thomas Funkhouser, 
Derek Hoiem, 
Alexei A. Efros, Andrew Owens, Antonio Torralba, Siddhartha Chaudhuri, and Szymon Rusinkiewicz for valuable discussion.

{\small
\bibliographystyle{ieee}
\bibliography{ShapeNet}
}

\end{document}